\documentclass[12pt, fceqn]{article}
\usepackage{amsfonts, amsthm, latexsym, amssymb}

\usepackage[fleqn]{amsmath}
\usepackage{graphicx}
\usepackage{graphics}
\usepackage[comma, numbers]{natbib} 
\usepackage[comma]{natbib} 
\usepackage{color}
\usepackage{comment}
\usepackage{algorithm}
\usepackage{algorithmic}
\usepackage{lscape}
\usepackage{footnote}
\usepackage{url}

\usepackage{epsfig}
\usepackage{fancybox}
\usepackage{psfrag}
\usepackage{tabularx}
\usepackage{float}
\usepackage{multirow}
\usepackage{rotating}
\usepackage{cancel}
\usepackage{epstopdf}
\usepackage{booktabs}

\usepackage[normalem]{ulem}

\date{}

\theoremstyle{remark}

\theoremstyle{definition}

{\begin{list}{}{\setlength{\itemsep}{4pt}
			\parindent 0pt  \setlength{\parsep}{0pt}\setlength{\leftmargin}{+25pt}
		\setlength{\itemindent}{-\parindent}}}{\end{list}}
\begin{document}
	
\begin{center}
	{\Large \textbf{Integrating AI and Ensemble Forecasting: Explainable Materials Planning with Scorecards and Trend Insights for a Large-Scale Manufacturer}}\\[12pt]
	
	
	\footnotesize
	
	\mbox{\large Saravanan Venkatachalam}\\
	
	Department of Industrial and Systems Engineering, Wayne State University, \\ 4815 Fourth St, Detroit, MI 48202.\\
	Corresponding author: \mbox{saravanan.v@wayne.edu}\\
	\normalsize
\end{center}

\begin{abstract}
This paper presents a practical architecture for after-sales demand forecasting and monitoring that unifies a revenue- and cluster-aware ensemble of statistical, machine learning, and deep learning models with a role-driven analytics layer for performance scorecards and trend diagnostics. The forecasting framework incorporates exogenous signals (installed base, pricing, macroeconomic indicators, life cycle, and seasonality) and explicitly treats COVID-19 as a distinct regime, producing country–part forecasts with calibrated intervals. A Pareto-aware segmentation models high-revenue items individually and pools long-tail items via clusters, while horizon-aware ensembling aligns weights with business-relevant loss functions (e.g., WMAPE). In addition to these forecasts, the performance scorecard provides decision-focused insights, including accuracy within tolerance thresholds by revenue share and count, bias decomposition (over- and under-forecast), geographic and product-family hotspots, and ranked root causes tied to high-impact part–country pairs. Complementing the static snapshot, the trend component evaluates trajectories of MAPE/WMAPE and bias across recent months, flags entities that are improving or deteriorating, detects change points aligned with known regimes, and attributes movements to lifecycle and seasonal factors. Critically, large language models (LLMs) and AI are embedded in the analytics layer to generate role-aware narrative outputs and enforce reporting contracts for the scorecards. They standardize business definitions, automate quality checks and reconciliations, and translate quantitative results into concise, explainable summaries for planners and executives. The system exposes a reproducible workflow—request specification, model execution, database-backed artifacts, and AI-generated narratives—so planners can move from “how accurate are we now?” to “where accuracy is moving and which levers to pull,” closing the loop between forecasts, monitoring, and inventory decisions across more than 90 countries and approximately 6,000 parts.
\end{abstract}



\section{Introduction}\label{sec-intro}

For any manufacturer, after-sales revenue is substantial—and in the automotive industry, it is pivotal. That revenue depends on the type and number of vehicles sold, as well as the position each product holds in its life cycle. The after-sales market delivers not only material income but also the parts availability that sustains high service levels and customer retention. On one hand, planners must understand the life-cycle phase to decide what to continue, phase down, or discontinue; on the other, they must uphold stringent service targets. Doing both well requires precise insight into future requirements.

In automotive specifically, spare parts demand behaves very differently from finished-vehicle demand. It is long-tailed, intermittent, and skewed: thousands of SKUs sell only a handful of units per year, while a small core turns constantly. Seasonality, weather, model launches and retirements, warranty cycles, recalls, supersession chains (where old part numbers are replaced by new), macroeconomic conditions, replenishment policies, tariffs, and other factors all influence the signal. Lead times vary widely across global suppliers, and service levels are unforgiving because a missing \$12 sensor can immobilize a \$60,000 car. A capable forecasting system acknowledges these realities: it detects level shifts after redesigns, respects obsolescence curves, and treats intermittent items differently from fast movers.

Accurate time-series forecasts do far more than juggle stock and cost. They underpin multi-echelon planning across plants, regional distribution centers, and dealer shelves, shape procurement contracts and expedite decisions, and inform pricing and repair-kit design by revealing attachment rates and failure patterns over time. As connected-vehicle, telematics, and warranty data mature, time-series models can fuse traditional history with leading indicators—such as miles driven, duty cycles, and diagnostic trouble codes—to anticipate failures before they occur. The benefits are tangible: higher fill rates and shorter backorders, lower working capital tied up in slow movers, fewer emergency expedites, and a better service-bay experience. In a market where brand trust depends on post-sale reliability, forecasting is not a back-office task—it is a strategic capability that keeps automobiles (and customers) moving.

Beyond the forecasts themselves, business users need clear market signals. They care about which markets are emerging, whether growth is consistent, how risks are distributed, how per-unit volume and revenue stack up, whether patterns are shifting month to month, and—across multiple regions—where sales are stable versus vulnerable. Meanwhile, algorithm and analytics teams focus on the consistency of error metrics over time, materials, and countries; on the SKUs that most influence safety stock; and on the interventions that will reduce forecast error. Both groups converge on the same outcomes: next steps and targeted recommendations to raise service levels while minimizing inventory.

This work presents a practical pipeline in which a forecasting algorithm and AI-driven analytics complement each other to deliver high-quality time-series forecasts for an automotive manufacturer operating in 90+ countries with over \$110M in revenue. \uline{The forecasting layer converts noisy historical demand into forward-looking signals with prediction intervals that guide production plans, stocking policies, and dealer replenishment—without flooding the network with inventory. The AI layer distills high-value insights from past data—trend and growth diagnostics, system performance, scorecards, and risk/opportunity highlights—so both business and technical users can act with confidence.}

\begin{figure}
	\centering
	\includegraphics[scale=0.35]{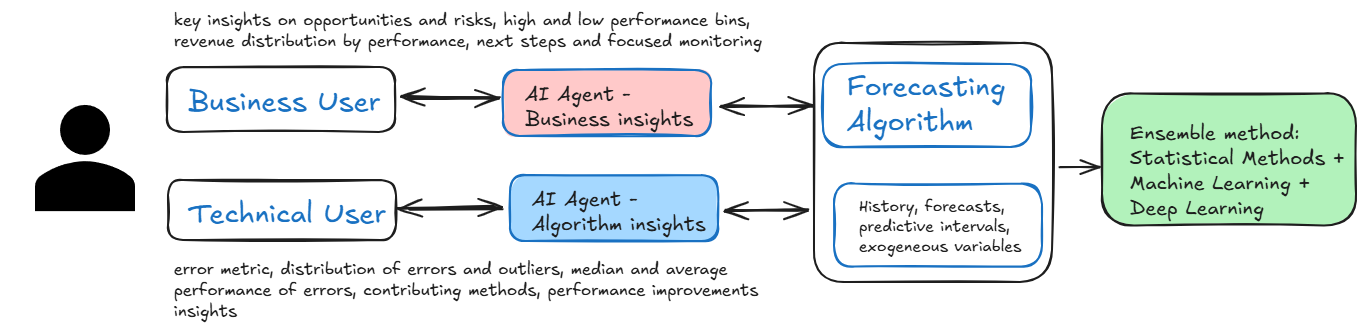}
	\caption{The diagram shows two AI agents—one for business insights and the other for algorithm insights—the agents are positioned between the users and the forecasting algorithm. The forecasting engine (an ensemble of statistical, ML, and deep-learning methods) ingests history and exogenous variables, and produces  forecasts and  predictive intervals. The business user AI agent provides perspectives on business actions (opportunities/risks, performance bins, revenue mix, and next steps), while the other AI agent offers insights on technical diagnostics (error metrics, distributions, drivers, and improvement ideas).}
	\label{fig:overall}
\end{figure}

Figure \ref{fig:overall} illustrates a two-lane decision flow: an ensemble ``Forecasting Algorithm'' (statistical + ML + deep learning) ingests history and exogenous drivers, then feeds two AI agents. The ``Algorithm Insights'' agent translates forecasts into technical diagnostics (error metrics, outliers, method contribution, improvement ideas) for ``Technical Users'', while the ``Business Insights'' agent turns the same outputs into actions (opportunities/risks, performance bins, revenue impact, next steps) for ``Business Users''. This separation of concerns makes the system a reusable framework—one source of forecast truth, two tailored views—so teams can both improve model quality and drive operational decisions from the same pipeline.

The diagram shows two AI agents—one for business insights and the other for algorithm insights—the agents are positioned between the users and the forecasting algorithm. The forecasting engine (an ensemble of statistical, ML, and deep-learning methods) ingests history and exogenous variables, and produces  forecasts and  predictive intervals. The business user AI agent provides perspectives on business actions (opportunities/risks, performance bins, revenue mix, and next steps), while the other AI agent offers insights on technical diagnostics (error metrics, distributions, drivers, and improvement ideas).

The diagram illustrates an end-to-end loop that transforms raw data into informed decisions, maintaining technical rigor and business impact in alignment. On the far right, an ensemble forecasting engine blends statistical baselines with machine-learning and deep-learning models. It consumes history, exogenous drivers, and constraints, and produces forecasts with predictive intervals. Those outputs don’t go straight to dashboards; they first pass through two specialized AI agents. The ``Algorithm-Insights'' agent speaks the language of data science—surfacing error metrics (MAPE, bias), error distributions and outliers, method contributions, stability over time, and concrete levers to improve the model (feature tweaks, segmentation, hierarchy pooling). The ``Business-Insights" agent translates the same forecasts into commercial insights—opportunities and risks by country/material, high/low performance bins, revenue at risk, next best actions, and focused monitoring rules.

On the left, two user groups consume tailored views without losing alignment. Technical users utilize the algorithm track to harden the system, diagnosing degradation, comparing model families, tuning parameters, and validating improvements against predictive intervals. Business users utilize the business track to plan inventory, pricing, promotions, and service levels, guided by clear prioritization (e.g., ``top bins to fix," ``segments driving most revenue variance"). Because both tracks are fed by the same forecasting truth, debates center on actions rather than reconciliations.

As a reusable framework, this flow scales to most real-world applications—such as retail, manufacturing, logistics, and finance—where forecasts need to be both trustworthy and actionable. It enforces a clean separation of concerns (science vs. strategy), bakes in governance (traceable metrics, auditable improvement ideas), and creates a continuous improvement loop: model outputs → diagnostic insights → targeted fixes and business actions → new data → better forecasts. The result is a reliable bridge from sophisticated algorithms to day-to-day decisions, accelerating time to value while maintaining transparency and control.

\section{Background and Motivation} \label{subsec:contributions}
In response to this need, we explore the integration of Large Language Models (LLMs) as a natural language interface to forecasting model engines. LLMs can interpret, contextualize, and summarize past data and forecasting scorecards while tailoring explanations to specific user roles. Their ability to dynamically generate multi-level, query-driven content enables real-time interactivity in ways that traditional dashboard tools cannot match. By embedding LLMs into the loop, we enable a form of ``explainable forecasting''—where outcomes are not only computed but also communicated in a transparent, responsive, and role-aware manner. This hybrid architecture redefines the role of AI in forecasting—not as a replacement for human decision-making, but as a facilitator of faster, clearer, and more informed decisions.

\subsection{Forecasting Methods} \label{subsec:optmodels}
Forecasting in manufacturing underpins production planning, procurement, inventory control, and after-sales service. Over two decades of empirical evidence demonstrate that no single model class dominates across the diverse, intermittently demanded series common in industrial contexts; instead, ensembles that combine statistical, machine-learning, and deep-learning methods tend to deliver the most accurate and reliable forecasts. This review synthesizes core approaches, with an emphasis on ensemble design for practical deployment in manufacturing and spare parts settings. Manufacturing demand forecasting is intrinsically a time-series problem: planners must anticipate level, seasonality, promotions, and structural breaks to synchronize capacity, materials, and service levels. Evidence from large-scale forecasting competitions and field deployments consistently shows that combining models yields more accurate and robust predictions than selecting a single winner \citep{timmermann2006forecast, makridakis2018m4, makridakis2022m5}. In production environments, ensembles also hedge model-selection risk, stabilize performance under regime shifts, and offer decision-useful uncertainty estimates (prediction intervals) for safety-stock and service-level targets.

Classical univariate methods---exponential smoothing (ETS) and ARIMA/seasonal ARIMA---remain strong baselines due to transparency, fast training, and robustness in data-scarce settings. For spare parts and maintenance, repair, and operations (MRO), intermittent demand is pervasive; Croston's method and its refinements (Syntetos--Boylan Approximation, TSB) explicitly model the occurrence and size of non-zero demand, reducing bias relative to naive approaches \citep{croston1972, syntetos2005sba, teunter2011tsb}. These models often serve as the backbone for slow-moving SKUs, with decomposition variants handling calendar and seasonal effects. Manufacturing data are hierarchical (plant $\rightarrow$ DC $\rightarrow$ dealer; item $\rightarrow$ family $\rightarrow$ category) and often grouped by geography and channel. Forecast reconciliation methods (e.g., MinT) adjust base forecasts to cohere across aggregation levels while minimizing variance \citep{wickramasuriya2019mint}. Coherent ensembles ensure that tactical plans at the leaf level add up to the strategic aggregates used for S\&OP.

Machine-learning (ML) models, such as gradient-boosted trees and random forests, are typically trained as global models across multiple related series with engineered features (e.g., calendar, promotions, prices, exogenous drivers). Their strength is capturing nonlinear interactions and leveraging rich covariates at scale. Results from the M5 competition (retail demand) highlight the competitive performance of boosted-tree families in high-dimensional, feature-rich settings \citep{makridakis2022m5}. In manufacturing, ML modules are effective components within ensembles, particularly when exogenous information — such as warranty claims, macroeconomic indicators, and telematics — is available.
Neural sequence models learn shared representations across thousands of series and capture long-range dependencies. Probabilistic RNN-based frameworks, such as DeepAR, provide calibrated distributions, enabling the direct optimization of likelihoods relevant to inventory decisions \citep{salinas2020deepar}. More recently, purely feed-forward architectures (e.g., N-BEATS) have matched or exceeded classical baselines on diverse benchmarks while preserving a degree of interpretability through basis expansions \citep{oreshkin2020nbeats}. Convolutional and Transformer variants broaden the design space further; in practice, they complement rather than replace statistical and ML methods.

Theoretical and empirical literatures agree that forecast combinations reduce variance and model-selection risk, delivering accuracy gains that persist across domains \citep{timmermann2006forecast}. The M4 competition showed that most top-performing entries were combinations or hybrids rather than single models \citep{makridakis2018m4}. Ensembles benefit from diversity across bias--variance profiles, feature sets, and inductive biases. Weighting schemes range from simple averages to stacking with cross-validated weights; in operations, simplicity and stability often trump marginal in-sample gains. Spare-parts demand is long-tailed and zero-inflated. For these series, intermittent-aware statistical models remain indispensable, while ML/DL components contribute via global structure, covariates, and meta-learning (e.g., learning when to prefer Croston, ETS, or N-BEATS). Practical ensembles treat intermittency as a first-class segmentation, with one branch specializing in slow movers (Croston/SBA/TSB), another covering medium to fast movers (ETS/ARIMA, global ML), and neural components capturing complex seasonality and cross-series patterns.

Evaluation should reflect operational costs: scale-free metrics (MAE/MASE), service-level considerations (pinball loss for quantiles), and intermittent-aware diagnostics. Out-of-sample backtesting with rolling origin evaluation is essential to avoid look-ahead bias. For deployment, key concerns include data latency, throughput, monitoring (including drift detection and stability of prediction intervals), and governance (traceability for S\&OP). Ensembles facilitate A/B testing and gradual rollout, enabling robust performance under policy or catalog changes.

A pragmatic blueprint for manufacturing is:
\begin{enumerate}
	\item Establish robust statistical baselines (ETS/ARIMA; Croston/SBA/TSB for intermittency) with hierarchical reconciliation.
	\item Add a global ML model to exploit exogenous features and cross-series sharing.
	\item Incorporate a neural component (e.g., DeepAR or N-BEATS) for complex patterns and probabilistic outputs.
	\item Combine via simple or stacked ensembles; calibrate prediction intervals.
	\item Reconcile across hierarchies to ensure plan coherence; monitor drift and rebalance weights over time.
\end{enumerate}
This ensemble-first strategy is evidence-based, operationally tractable, and resilient to regime shifts typical in manufacturing.

\subsection{Large Language Models} \label{llms}
Large language models (LLMs) are playing a significant role in the growing application of AI across various domains \cite{github2023copilot, chen2023frugalgpt}. In natural language processing, LLMs have replaced traditional statistical and rule-based methods with neural networks trained on vast amounts of text data, enabling them to capture complex linguistic patterns and relationships \cite{devlin2018bert, rosset2020turing-nlg, smith2022using}. Built on transformer architectures that combine encoders and decoders, LLMs efficiently process sequences of text based on learned context \cite{liu2023prompt, brown2020language}. They are increasingly impactful in data analysis, where natural language queries can be translated into executable code for analysis and visualization, allowing business users to perform complex tasks with ease \cite{danilevsky2020survey, ahmed2022artificial}. Moreover, LLMs excel at interpreting results by transforming the outputs of OR models—such as solution values, dual variables, and slackness conditions—into clear, coherent explanations for non-technical users. This automated data storytelling enhances understanding and supports better decision-making \cite{bommasani2021opportunities, peters2018deep}. Integrating LLMs into OR model pipelines thus improves both the execution and interpretation of results, making insights more accessible and boosting confidence among business users.

Additionally, users within an organization often have varying informational needs based on their roles. For example, analysts may be interested in insights at the item level, while managers focus on product families, and senior executives are typically concerned with performance at the regional or location level. Designing static user interfaces to cater to each of these perspectives can be both complex and inflexible. However, with the capabilities of LLMs, dynamic and role-specific descriptive summaries of key performance indicators (KPIs) can be generated on demand. This adaptability allows each user to receive insights tailored to their level of responsibility and decision-making needs. As a result, integrating OR models with LLMs creates a comprehensive and flexible pipeline that supports users across different organizational levels, improving accessibility, clarity, and overall effectiveness in decision-making.

\section{Description} \label{sec:formulation} 

The OEM manages two broad automobile families (about 25 models each) and roughly 6,000 spare parts across more than 90 countries, and the planning goal is to forecast country–part demand and set stock per country without splitting demand, acknowledging that demand is shaped by installed base and model mix, price bands, macroeconomic conditions, life-cycle phase, and seasonality, with curated exogenous data on models and processes and explicit treatment of COVID-19 as a special regime rather than noise. Because revenue is highly skewed (approximately 80\% from <10\% of parts), we segment series into a high-revenue tier modeled individually and a long-tail tier handled by clusters that share statistical strength via global models; clusters are formed with descriptors like intermittency (ADI), dispersion (CV$^2$), seasonality strength, price band, and life-cycle state. The forecasting engine is an extensible ensemble that blends three streams—classical statistical methods (e.g., exponential smoothing, SARIMA with regressors), machine learning (e.g., gradient-boosted trees on lags, calendars, and covariates), and deep learning sequence models—using non-negative horizon-aware weights learned by rolling-origin validation to minimize a business-aligned loss such as WMAPE, with optional meta-learning so weights depend on horizon, revenue tier, and cluster features; if one stream (say ML) consistently dominates, the registry promotes more models from that stream without refactoring. COVID effects enter via regime flags (shock/restriction/recovery), change-point features, and robust losses (Huber/quantile), while statistical models allow intervention dummies and ML/DL ingest regime and time-since-regime covariates to capture recovery. Training and governance follow a disciplined MLOps loop: feature engineering for economics, lifecycle, calendars, and holidays; rolling backtests with latency-aware gaps; tracking of MAPE/WMAPE and service-oriented error percentiles for safety-stock sizing; a model registry that records candidates, scores, and ensemble weights; and drift monitors on covariates and residuals that trigger re-tuning or re-weighting. Outputs include point forecasts and calibrated intervals delivered at the country-part level to downstream inventory, where the “no demand splitting” constraint is enforced. The overall architecture is deliberately modular, revenue-focused, and cluster-aware, allowing it to absorb better models over time while remaining robust across regimes and horizons.

\section{Technical Architecture} \label{ta}

\begin{figure}[H]
	\centering
	\includegraphics[scale=0.5]{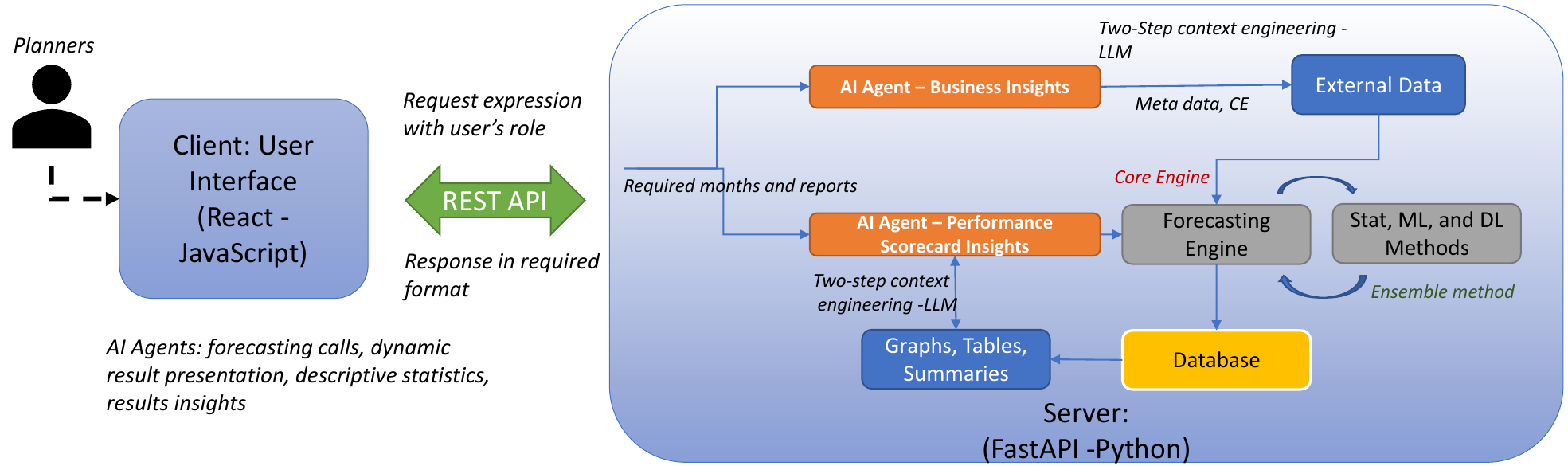}
	\caption{System architecture for descriptive business insights and forecasting  with AI agents, optimization engines, and dynamic result generation.}
	\label{fig:sysarch}
\end{figure}

The illustrated architecture in Figure \ref{fig:sysarch} presents an AI-powered decision-support system tailored for insights on operations and forecasting in supply chain planning. It is designed to support planners across various roles by integrating a user-friendly interface with a powerful backend system. The backend consists of AI agents, a forecasting engine, and a dynamic trend interpretation layer. The system emphasizes interactivity, role-aware recommendations, and explainability—bridging the gap between complex forecasting models and real-world decision-making.

\noindent Figure~\ref{fig:sysarch} depicts the end-to-end workflow that operationalizes forecasting and insight generation for after-sales parts planning. Planners interact through a role-aware client implemented in React/JavaScript, which standardizes user intents and converts them into structured requests specifying horizons, geographies, product scopes, and report types. These requests are sent to the server through a REST API that enforces authentication and authorization and ensures a canonical schema for downstream components, thereby supporting reproducibility and audit trails. Inside the FastAPI/Python server, two specialized language-model agents orchestrate context and reporting using a two-step prompt-engineering procedure. The business insights agent fuses request metadata with external signals such as macroeconomic indicators, calendar effects, product life-cycle flags, and extraordinary regimes like COVID-19 to construct machine-readable context for modeling. In parallel, the performance scorecard insights agent aligns evaluation requirements with organizational scorecards, prepares metric definitions and report windows, and later translates raw outputs into narratives, tables, and figures tailored to user roles. The forecasting engine forms the computational core and exposes a uniform interface across classical statistical models, machine-learning algorithms, and deep-learning sequence learners. Base forecasts are combined by an ensemble layer with nonnegative, horizon-aware weights that are learned via rolling-origin validation against business-aligned losses, and the engine emits both point and interval forecasts at the country–part level together with diagnostic metadata. A database underpins both training and reporting by supplying curated inputs, capturing forecasts and residuals, and maintaining versioned artifacts for governance, replayability, and error analysis. The reporting path queries this store to assemble graphs, tables, and summaries, including revenue-weighted aggregates and cluster-level breakouts, which are returned through the API to the client for interactive exploration. Overall, Figure~\ref{fig:sysarch} emphasizes clean separation of concerns: intent capture and access control in the client and API, context construction and narrative generation via LLM agents, robust and extensible multi-model forecasting in the core engine, and durable persistence that guarantees traceable, reproducible analytics suitable for enterprise deployment and scholarly evaluation.

\section{Implementation} \label{sec:algorithms}

This section describes the end-to-end implementation of the forecasting and context engineering system. The goal is to deliver explainable, data-driven insights to supply chain planners and decision-makers. The implementation integrates a dynamic optimization dashboard, real-time transfer visualization, scorecard tracking, and a robust context engineering framework powered by LLMs. The system architecture supports operational responsiveness and explainability through an interactive user interface and intelligent backend reasoning.

\subsection{Dashboard}

\begin{figure}[H]
	\centering
	\includegraphics[width=\textwidth]{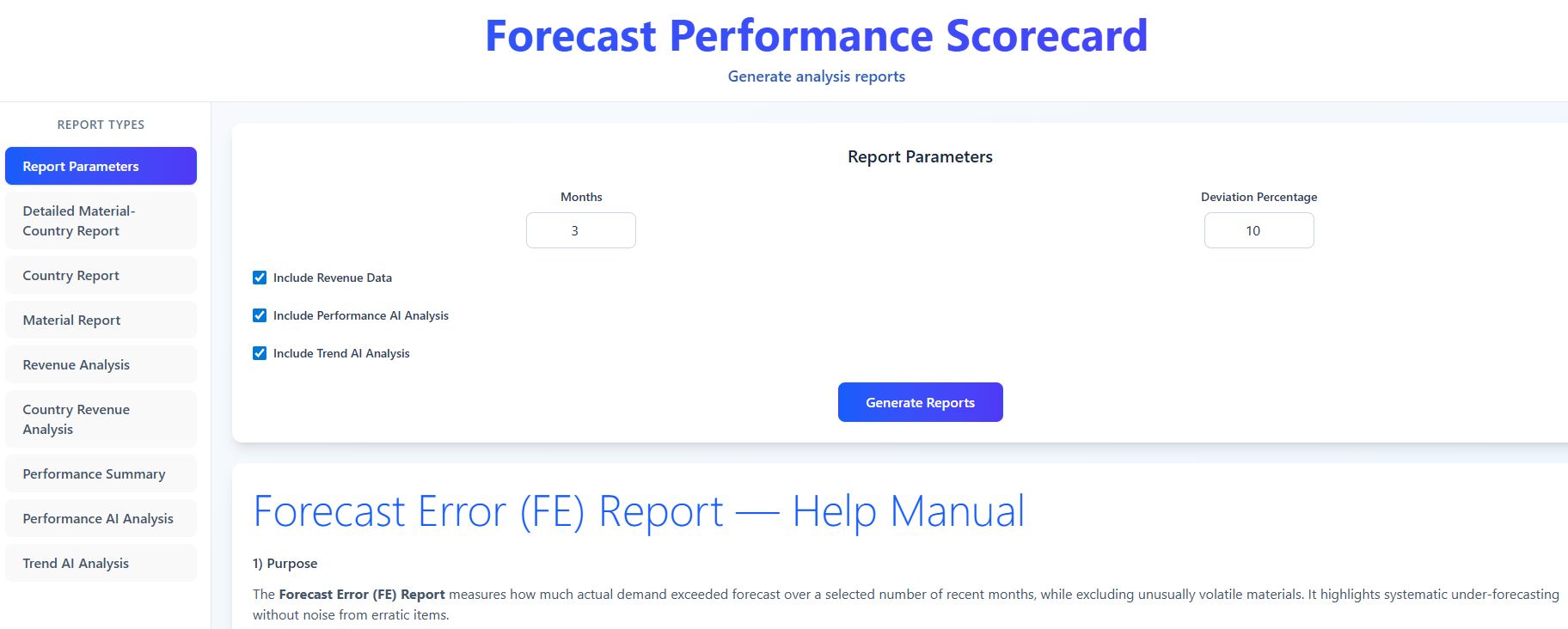}
	\caption{Forecast Performance Dashboard interface displaying real-time metrics on various metrics, insights, and recommendations for next steps. The dashboard enables users to get insights on performance and trend for materials planning related to forecasting and safety stocks.}
	\label{fig:dash}
\end{figure}

Figure \ref{fig:dash} illustrates the control interface for the Forecast Performance Scorecard, which operationalizes evaluation and monitoring of the ensemble forecasting framework described earlier. User intents are specified through two primary parameters: the number of recent months over which evaluation is performed and a deviation percentage that defines the band for acceptable forecast error calculations. These parameters are complemented by three toggles that determine whether revenue-weighted views, automated performance narratives, and automated trend diagnostics are included in the generated report. The design encourages reproducible, role-consistent analysis by translating free-form planner actions into a canonical specification that is passed to the server-side analytics services.

The performance scorecard summarizes accuracy over the selected horizon with an emphasis on economically material outcomes. Because after-sales demand exhibits a strong Pareto structure—where fewer than ten percent of parts generate approximately eighty percent of revenue—the scorecard privileges revenue-weighted measures such as WMAPE alongside unweighted MAPE. In practice, this means that even when a minority of series fall within the specified tolerance, the majority of revenue may still be well served; conversely, small counts of large deviations can dominate the business impact. The scorecard decomposes these metrics by country, product family, and revenue tier, reflecting the architecture’s segmentation into individually modeled high-revenue items and cluster-modeled long-tail items. It also reports bias directionality to distinguish under-forecasting, which risks service failures, from over-forecasting, which inflates working capital. These views are tied to versioned artifacts in the data store, ensuring traceability of each figure and table to a specific data slice and model registry state.

Trend analysis augments the static performance snapshot with a lightweight time-series treatment of evaluation metrics themselves. Over the user-selected window, the system estimates slopes and detects change points for MAPE/WMAPE and bias, classifying each entity's trajectory as improving, deteriorating, or stable. The trend agent integrates exogenous context—such as product life-cycle flags, seasonality, and COVID-period indicators—so that level shifts aligned with known regimes are attributed appropriately rather than mischaracterized as model drift. For the cluster-modeled long tail, trend diagnostics are also aggregated by descriptors of intermittency and dispersion, which helps distinguish issues that call for policy interventions (e.g., safety-stock rules for intermittent demand) from issues that warrant re-weighting or re-training of forecasting models.

The scorecard's insights are explicitly connected to the ensemble architecture. Because model weights are learned via rolling-origin validation and are horizon- and tier-aware, the performance views can indicate which model stream—statistical, machine learning, or deep learning—dominated the ensemble in the evaluation window. When systematic patterns emerge (for example, machine-learning models consistently outperforming at short horizons for high-revenue parts while statistical baselines remain competitive for highly intermittent items), these observations inform the registry's promotion policy and subsequent re-allocation of modeling effort. In this way, the user interface does not merely report accuracy; it closes the loop between monitoring and governance by surfacing evidence that supports principled adjustments to the modeling portfolio.

From an operational standpoint, the interface encourages disciplined use. Short windows with moderate tolerance thresholds provide timely feedback for planners, while longer windows stabilize trend classification when recent volatility is suspected. Enabling revenue views focuses attention on what matters economically, and enabling the automated narratives produces decision-ready summaries for stakeholders who do not need to inspect series-level residuals. Together, the performance and trend components of the scorecard turn planner inputs into a coherent narrative about current accuracy, directional movement, and probable levers—whether to adjust ensemble weights and covariates for specific high-impact cohorts or to review inventory policies for structurally intermittent clusters.

\subsection{Context Engineering Framework}

\begin{figure}[H]
	\centering
	\includegraphics[width=\textwidth]{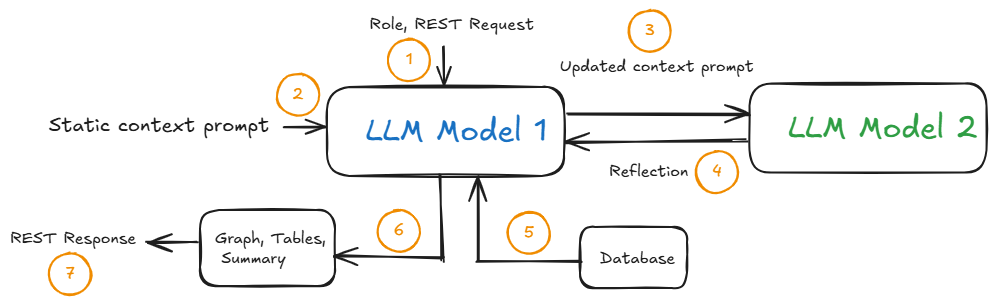}
	\caption{Context engineering architecture integrating LLMs, REST APIs, and data systems to deliver interactive, explainable, and role-specific decision support in seven structured steps.}
	\label{fig:ce-arch}
\end{figure}

\noindent Figure~\ref{fig:ce-arch} summarizes the two-model context-engineering (CE) loop that turns free-form planner intents into specification-driven analyses for the performance scorecard and trend reports. An authenticated REST request, tagged with the user’s role and desired report type, first enters LLM~Model~1 (1), which acts as the orchestrator by normalizing the request into a canonical job specification (scope, months, thresholds, report family). Model~1 is primed by a static, task-specific context prompt (2) drawn from the attached context-engineering contracts: the performance scorecard specification for error and revenue-weighted diagnostics, the trend specification for growth-mix structure, and the monthly trend specification for short-horizon dynamics. Using these contracts, Model~1 composes an updated, executable context prompt (3) that resolves datasets, filters, metrics, required tables and figures, and quality checks, and then dispatches this prompt to LLM~Model~2, the task specialist. Model~2 executes the analysis and returns a structured reflection to Model~1 (4) containing computed KPIs (e.g., MAPE/WMAPE, bias, revenue shares), ranked country/material tables, bin distributions, change-point and slope assessments, as well as diagnostics if any specification constraints are violated (for example, reconciliation of totals, bin completeness, or denominator validity). Where necessary, Model~1 mediates access to the database layer (5) to read curated inputs and persist versioned outputs, thereby ensuring lineage, replayability, and auditability across runs. Model~1 then assembles publication-ready artifacts (6)—graphs, tables, and an executive summary—strictly following the formatting and interpretation rules stipulated in the contracts (for instance, one chart per figure with matplotlib, spreadsheet-style tables, explicit revenue weighting, and role-aligned narratives). Finally, the server returns a deterministic REST response (7) that bundles the narrative, figures, and tables; because the contracts govern both computation and presentation, repeated requests with the same parameters produce consistent results, while still allowing planners to switch seamlessly between scorecard accuracy views and trend-oriented insights. In combination, the orchestrator–specialist split enforces a clean separation of concerns: Model~1 guarantees contextual fidelity to the performance and trend specifications, and Model~2 concentrates on domain calculations and validation, yielding analyses that are both explainable and operationally robust.

\subsection{Performance Analysis using AI}

The CE equips users to translate raw forecast outputs into clear, business-ready insights. It centers its analysis on mean absolute percentage error (MAPE) to quantify how forecasts perform overall and within key segments. Users are guided to produce an executive summary that answers the following questions: what the typical error looks like (mean and median), how it shifts under a filtered lens versus all items, and where the largest, most actionable gaps reside. By providing a concise yet comprehensive overview, decision-makers can quickly assess whether performance is broadly acceptable or if there are pockets of high error that warrant targeted intervention.

A critical contribution of the brief is its explicit treatment of filtered versus unfiltered views. By instructing analysts to treat the filtered view as primary, while using the all-items summary only for short context, the CE avoids diluting the signal with extreme outliers or noisy tails. This framing enhances comparability over time and across segments, and helps stakeholders distinguish between genuine improvement and mechanical effects. Furthermore, the inclusion of revenue-weighted MAPE enables prioritization by business impact: a modest error in a large-revenue segment can outrank a large error in a small-revenue niche.

The context also formalizes how to communicate forecast accuracy distribution via MAPE bands. By reporting the share of materials in each band (\(<10\%\), \(10\text{--}20\%\), \(20\text{--}40\%\), \(>40\%\)), users obtain an intuitive picture of breadth versus concentration of error. If most materials have an MAPE below \(20\%\) but a non-trivial tail exceeds \(40\%\), the organization can target remediation where it matters. Extending the same view to countries or size classes further pinpoints whether high-error items are isolated, clustered in specific geographies, or concentrated among large materials that drive disproportionate revenue.

Material-level and country-level sections translate accuracy into actions. For materials, the brief recommends surfacing both high-revenue, high-MAPE outliers (risk) and high-revenue, low-MAPE items (wins), encouraging teams to preserve what works and remediate what does not. For countries, computing mean, median, and revenue-weighted MAPE—alongside country revenue—reveals where accuracy problems impose the greatest commercial cost. Ranking top and bottom performers and flagging “high revenue and high MAPE” countries makes it straightforward to align inventory, pricing, or promotion strategies with the true error landscape.

Revenue-bin analysis adds another diagnostic layer by examining \(\text{avg\_mape}\) across bins and the underlying \(\text{material\_count}\). This helps determine whether accuracy systematically improves with scale or degrades in specific value bands, and which bins hold enough items to justify model specialization. A deviation analysis by country (where a country’s bin pattern diverges from the overall trend) highlights where local market dynamics, data quality, or modeling assumptions may require bespoke treatment. Together, these views link statistical performance to commercial structure, improving both model governance and business planning.

Finally, the CE codifies clear deliverables and quality checks that raise confidence in the results. Required tables and charts ensure that narrative claims align with evidence, while guardrails such as consistent sorting, one chart per figure, and restrained formatting keep the message legible. The quality checklist (e.g., ensuring that rankings match visuals, that filtered claims align with the specified summary, and that named outliers are present in the detailed file) helps prevent common errors in communication. In practice, this yields a repeatable workflow: calculate headline metrics, locate high-impact accuracy gaps by country and material, verify patterns in revenue bins, and close with prioritized recommendations tied to measurable improvements in forecast quality and business outcomes.

\subsection{Trend Analysis using AI}
\subsubsection{Trend Analysis using AI - Overall}
The CE establishes a clear blueprint for understanding where overall impact and opportunities reside when data are aggregated without an explicit time axis. Users will learn to quantify total actuals and revenue, compute shares, and evaluate concentration using tools such as the Pareto view and the Herfindahl–Hirschman Index. This matters in practice because leadership needs to know whether performance is broad-based or dominated by a few countries or materials. For example, discovering that the top three countries hold 62\% of revenue prompts targeted risk mitigation if any one of them stalls.

A central insight is the separation of volume and value through the average selling price, \(\mathrm{ASP} = \frac{R}{A}\), revenue (R) divided by quantity (A), and the premium/discount signal, \(\mathrm{PremDisc} = \mathrm{Share}_R - \mathrm{Share}_A\). Users can identify value-heavy segments (\(\mathrm{PremDisc} > 0\)) where pricing or mix is strong, and volume-heavy segments (\(\mathrm{PremDisc} < 0\)) where pricing headroom or mix upgrades may exist. In practical terms, a country contributing 9\% of revenue but only 5\% of actuals suggests a premium pocket worth protecting, whereas a material at 7\% of actuals but 3\% of revenue points to price or mix actions.

Another key stream of insight comes from the MAPE distribution across forecast-accuracy bins \textless 10\%, 10–20\%, 20–40\%, and \textgreater 40\%. By reporting the share of materials in each bin overall, by country, and by size class (Large/Medium/Small), the framework identifies where high error clusters are located and where they have the greatest impact on the business. For example, if Large materials show 28\% of items in the \textgreater 40\% bin while Small materials show only 12\%, the guidance is to prioritize model and data fixes on the high-impact Large segment. Weighted views (revenue- or actuals-weighted bin shares) help highlight the financial relevance of accuracy hotspots, while noting that these weights are heuristic and not a substitute for rigorous accuracy metrics.

Country- and material-level ranking tables convert diagnostics into direction. Users will see \(A\), \(R\), \(\mathrm{Share}_A\), \(\mathrm{Share}_R\), \(\mathrm{ASP}\), and \(\mathrm{PremDisc}\) side by side, making it straightforward to spot value-heavy countries with weak ASP relative to peers or materials whose low ASP depresses revenue share despite substantial volume. A concrete example is a brake component with high actuals but a low ASP, pulling down the value; the recommended action could involve combining pricing tests with a packaging change (e.g., service kits) to lift perceived value.

Intersection analysis (country \(\times\) material) reveals where the portfolio is over- or under-penetrated. A heatmap-like table of revenue or actuals by intersection, annotated with each intersection’s MAPE-bin profile, guides interventions to the exact places where commercial potential and forecast risk coincide. If a high-revenue country is underrepresented in a premium material and the material’s MAPE primarily falls within 20–40\% bins, users can target both go-to-market expansion and model refinement for that intersection.

The CE also embeds practical governance and data-quality safeguards that build trust. It instructs users to reconcile totals across files, handle undefined ASP when \(A=0\), winsorize ASP when summarizing outliers, and exclude undefined MAPE cases from bin percentages while reporting coverage (e.g., “MAPE available for 93\% of materials”). These steps prevent misleading conclusions and ensure that recommendations rest on stable calculations.

Ultimately, the CE prompts users to take action beyond mere description. It asks for prioritized recommendations tied to the observed mix, concentration, and accuracy patterns, such as improving data for countries with elevated \textgreater 40\% MAPE share, pricing or assortment moves for volume-heavy/low-ASP pockets, or monitoring rules that trigger alerts when high-impact segments show worsening accuracy. In day-to-day terms, this means planners know which SKUs to optimize for accuracy, commercial teams know where price or mix adjustments can enhance value, and leadership gains a clear, auditable view of where the business is strong, fragile, or underdeveloped.

\subsubsection{Trend Analysis using AI - Monthwise}
The CE brief equips users to extract clear month-over-month (MoM) signals from operational data and convert them into actionable decisions. It emphasizes a succinct executive summary that answers what changed last month, by how much, and why. Users obtain headline MoM shifts in actuals (the primary volume signal) and in revenue (the value lens), alongside the top positive and negative contributors across countries and materials. This enables a general manager or supply lead to see whether growth is broad-based or concentrated in a few segments, and where risk may be building, such as two consecutive negative months in a major market.

For country managers, the framework ranks contribution to the latest change in actuals, not just percent growth. This distinction is practical: a small market up 50 percent may add less volume than a large market up 2 percent. The deliverable includes a companion table showing last month versus prior (\(A_{t}, A_{t-1}\)), the absolute delta (\(\Delta A_t\)), and percent change where valid, helping separate real momentum from noise. For example, if Mexico is \(+8{,}200\) units MoM while Germany is \(-3{,}100\), the net direction is clear, and it becomes straightforward to identify the three countries that explain most of the total swing.

For product owners and inventory planners, the material analysis highlights top and bottom movers by absolute MoM change (what truly moved the needle) and by percentage MoM (where demand is accelerating or decaying fastest). It also spotlights large materials (for instance, the top 20 percent by trailing three-month size) that are shrinking or surging, since these have the highest impact. In practice, this might surface a brake-pad stock-keeping unit that dropped 12{,}000 units MoM despite being a top-revenue item, prompting checks on pricing, availability, or part supersession.

A further insight stream concerns price and mix diagnostics: by juxtaposing revenue with actuals, the brief separates volume effects from pricing or average selling price effects. Users see where revenue increases while actuals decrease (due to price or mix tailwinds) and the reverse (indicating potential price pressure or mix deterioration). This enables targeted levers, such as holding the price where the average selling price supports revenue, or fixing availability where actuals are up but revenue underperforms due to discounting.

The methodology incorporates data safeguards that are relevant in real-world settings. It avoids misleading percentages when the prior month is zero or missing by reporting not available for percent change and using absolute deltas. It discourages fabricating rows for gaps and promotes simple momentum checks (\(\Delta A_t - \Delta A_{t-1}\)) so teams can distinguish a one-off spike from a sustained trend. It also recommends governance checks, such as ensuring that country and material contributions reconcile to the total change, so leaders can trust that calculations are internally consistent before committing inventory or adjusting targets.

Finally, the brief pushes actionability. It guides users to produce a short list of prioritized recommendations and lightweight monitoring rules, for example, to alert when any top-size material posts two consecutive negative MoM actuals or an average selling price swing beyond a set threshold. In practice, this supports faster root-cause reviews, tighter sales and operations planning cycles, and fewer surprises: planners know which items to expedite, sales sees where to focus effort, and finance understands why revenue moved as it did.

\label{sec:comp} 
\section{Conclusion and Future Directions}\label{sec:con} 

This article presented a practical, production-grade framework that couples a revenue- and cluster-aware ensemble of statistical, machine-learning, and deep-learning forecasters with a two-step, role-aware context engineering layer. In a large, long-tailed after-sales setting spanning thousands of parts and dozens of countries, the approach delivers calibrated forecasts, interpretable scorecards, and trajectory-aware trend insights that map directly to inventory and service decisions. The architectural separation of concerns—forecast generation, explanation, and governance—proved essential for auditability, reproducibility, and rapid iteration when demand regimes and catalog composition shift.

A central contribution is the integration of private large language models as the interface between complex analytics and operational users. Deployed on-premise or within a virtual private cloud, private LLMs preserve data residency and confidentiality while enabling dynamic, role-specific narratives, quality checks, and specification enforcement for reports. This design converts free-form planner intent into deterministic, contract-driven analyses, returning evidence-linked narratives (including figures, tables, and summaries) that are consistent across runs and stakeholders. In practice, private LLMs function as “context engineers” that standardize business definitions (e.g., MAPE/WMAPE, revenue weighting, intermittency) and ensure that outputs remain traceable to the underlying data slices and model versions.

Equally important is the use of transfer learning \cite{weiss2016survey,lester2021power,dettmers2023qlora} to adapt both forecasting models and language interfaces to heterogeneous markets and product cohorts. On the forecasting side, global models exhibit statistical strength across related series, while parameter-efficient techniques (e.g., prompt tuning or low-rank adapters) enable rapid specialization to new countries, life-cycle phases, and pricing regimes without requiring retraining from scratch. On the language side, lightweight fine-tuning of private LLMs \cite{touvron2023llama} on organization-specific terminology, report templates, and error taxonomies yields sharper, more reliable explanations and reduces hallucinatory risk. Together, these transfer mechanisms shorten time-to-value, stabilize performance under regime change, and lower compute and data requirements for continuous improvement.

The resulting system creates a closed loop from forecasts to decisions: ensemble outputs feed private-LLM agents that diagnose performance, attribute drivers, and prioritize actions; those actions update data and policies that flow back into re-weighted ensembles and adapted prompts. This loop supports principled governance (versioned artifacts, drift monitoring, and lineage), while remaining extensible: additional model streams or exogenous data sources can be added without breaking client contracts, and private LLMs can be re-tuned incrementally as business vocabulary and objectives evolve. In summary, combining private LLMs with transfer learning within a modular ensemble-and-context architecture transforms forecasting from a point solution into an adaptive capability—secure, explainable, and economically aligned with the realities of after-sales planning.

\bibliographystyle{elsarticle-num} 
\bibliography{asd_fe}
\end{document}